\documentclass[letterpaper, 10 pt, conference]{ieeeconf}  

\IEEEoverridecommandlockouts                              

\let\labelindent\relax
\usepackage{times}
\usepackage{epsfig}
\usepackage{graphicx}
\usepackage{amsmath}
\usepackage{amssymb}
\usepackage{enumitem}
\usepackage{booktabs}
\usepackage{xcolor}
\usepackage{multirow}
\usepackage{subcaption}
\makeatletter
\let\NAT@parse\undefined
\makeatother
\usepackage[pagebackref=true,breaklinks=true,letterpaper=true,colorlinks,citecolor=blue,linkcolor=blue,bookmarks=false]{hyperref}

\usepackage[colorinlistoftodos,textwidth=\marginparwidth]{todonotes}

\title{\LARGE \bf
 Learning 3D-aware Egocentric Spatial-Temporal Interaction via \\Graph Convolutional Networks}

\author{Chengxi Li$^{1,3}$ \, Yue Meng$^{2}$ \, Stanley H. Chan$^{1}$ \, Yi-Ting Chen$^{3}$
\thanks{C. Li$^{1}$ and S. H. Chan$^{1}$ are with Department of Electrical and Computer Engineering, Purdue University, West Lafayette, IN, USA. Y. Meng$^{2}$ is with IBM Thomas J. Watson Research Center, Yorktown Heights, NY, USA. Chengxi Li$^{3}$ and Y.-T. Chen$^{3}$ is with Honda Research Institute USA, San Jose, CA, USA.}%
\thanks{The work was done when C. Li and Y. Meng were interns at Honda Research Institute, USA.}
}

\begin{document}

\maketitle
\thispagestyle{empty}
\pagestyle{empty}

\begin{abstract}
To enable intelligent automated driving systems, a promising strategy is to understand how human drives and interacts with road users in complicated driving situations. 
In this paper, we propose a 3D-aware egocentric spatial-temporal interaction framework for automated driving applications.
Graph convolution networks (GCN) is devised for interaction modeling.
We introduce three novel concepts into GCN.
First, we decompose egocentric interactions into ego-thing and ego-stuff interaction, modeled by two GCNs.  
In both GCNs, ego nodes are introduced to encode the interaction between thing objects (e.g., car and pedestrian), and interaction between stuff objects (e.g., lane marking and traffic light).
%
Second, objects' 3D locations are explicitly incorporated into GCN to better model egocentric interactions.
Third, to implement ego-stuff interaction in GCN, we propose a MaskAlign operation to extract features for irregular objects.

We validate the proposed framework on tactical driver behavior recognition.
Extensive experiments are conducted using Honda Research Institute Driving Dataset, the largest dataset with diverse tactical driver behavior annotations.
Our framework demonstrates substantial performance boost over baselines on the two experimental settings by 3.9\% and 6.0\%, respectively.
%
Furthermore, we visualize the learned affinity matrices, which encode ego-thing and ego-stuff interactions, to showcase the proposed framework can capture interactions effectively.         
\end{abstract}

\section{INTRODUCTION}
Automated driving in highly interactive scenarios is challenging as it involves different levels of 3D scene analysis~\cite{BarsanICRA2018,WangIV2018}, situation understanding~\cite{SchmidtITSC2014,Li_situatino_iccv2017}, intention prediction~\cite{LeeCVPR2017,Chandra_Traphic_CVPR2019}, decision making and planning~\cite{ZhanITSC2016,QiICRA2018}. 
%
%
Understanding how human drives and interacts with road users is essential toward an intelligent automated driving system.
The first step to achieve this is to develop a computational model which can capture the complicated spatial-temporal interactions between the ego vehicle and road users.

\begin{figure}[t!]
\vspace{8pt}
\minipage{\columnwidth}
  \includegraphics[width=\linewidth]{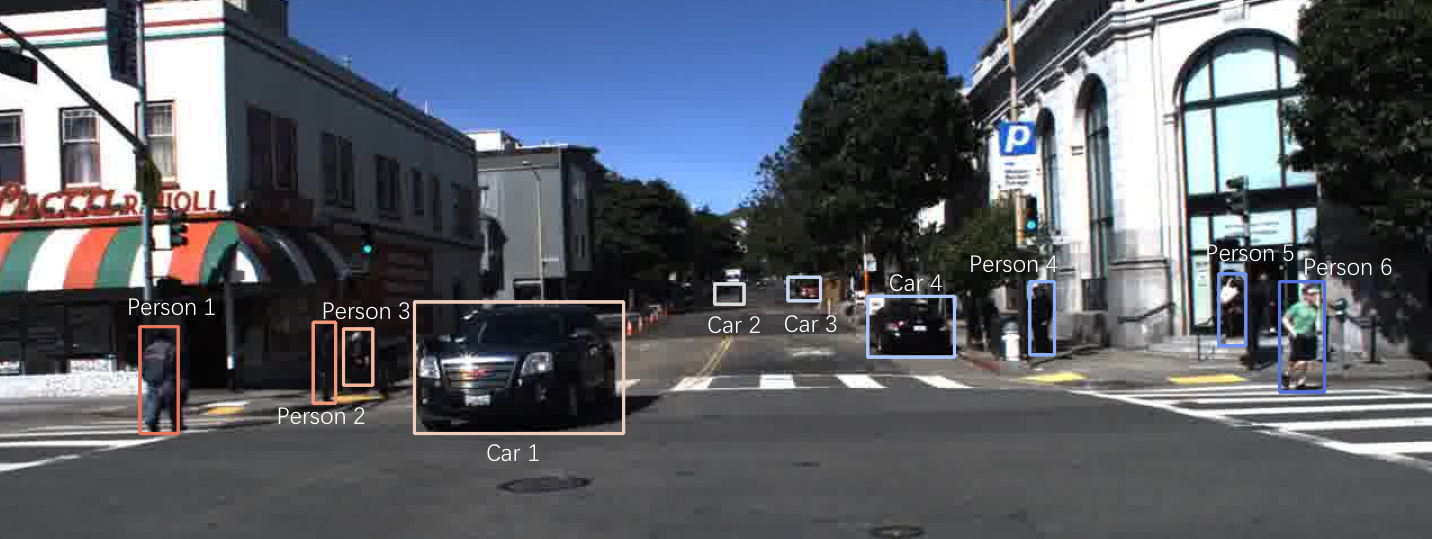}
  \vspace{-15pt}
  \caption*{ \footnotesize (a) \textbf{Goal-oriented} prediction: Left turn and \textbf{Cause} prediction: crossing vehicle.}
\endminipage \hfill

\minipage{0.38\columnwidth}
  \includegraphics[width=\linewidth]{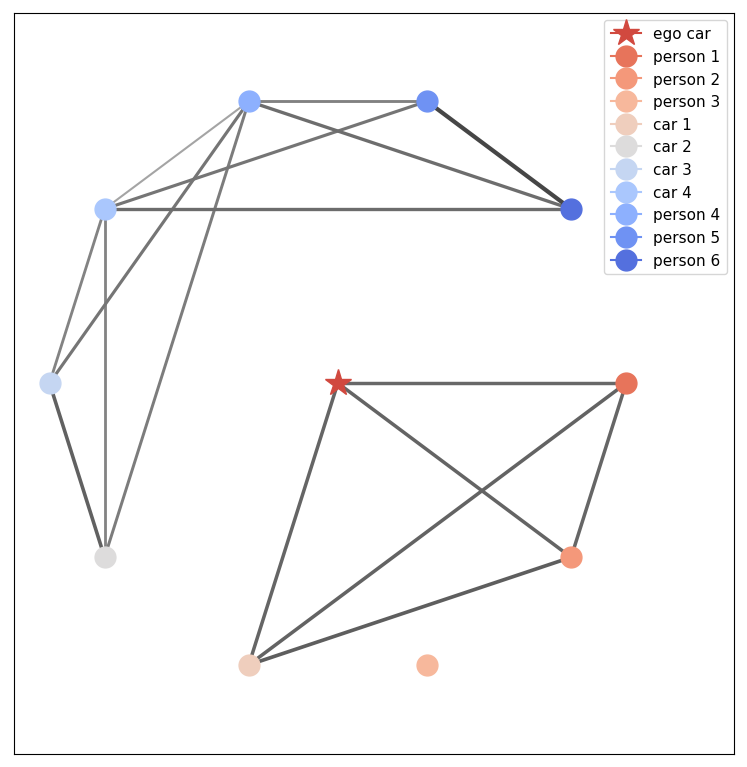}
  \caption*{\footnotesize(b) Learned affinity matrix }
\endminipage\hfill
\minipage{0.62\columnwidth}
 \vspace{8pt}
  \includegraphics[width=\linewidth]{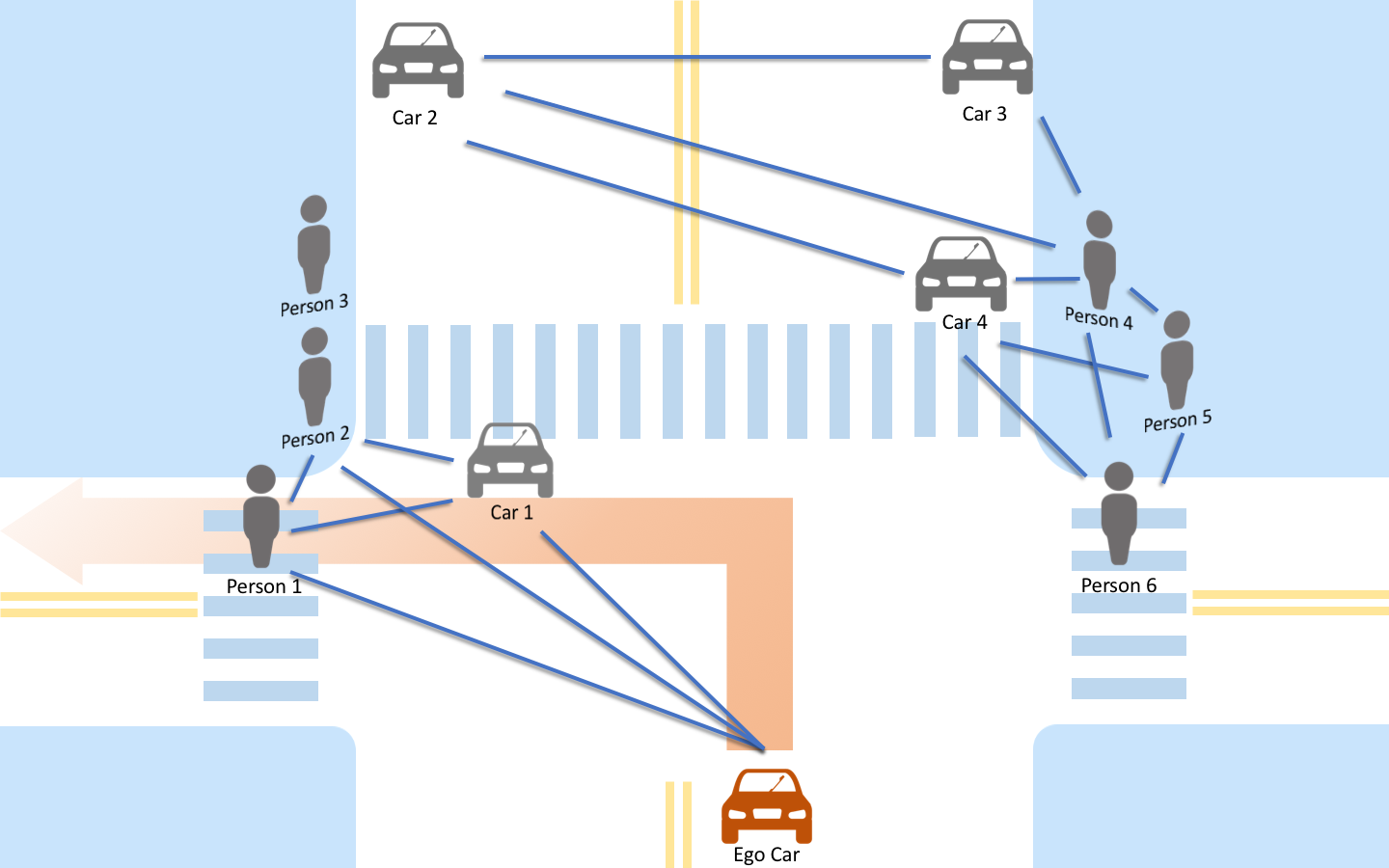}
  \caption*{\footnotesize(c) Visualization of the learned affinity matrix using a top-view scene layout}
  \endminipage
 \caption{In a complicated traffic situation at intersections, the ego vehicle intends to take a left turn while yielding to a upcoming vehicle. Our model learns a graph structure in (b) using edge connections to represent the interactions among road users and the ego vehicle. The top-view scene representation in (c) is derived from (a) and (b) by overlaying learned relations on a scene layout for better illustration.}
 \label{fig:visualization_scene_representation}
 \vspace{-10pt}
\end{figure}

%
Over the past decade there has been a significant advance in modeling spatial-temporal interactions~\cite{OliverIV2000,Mitrovic_ITS_2005,Jain_anti_maneuver_iccv2015,Gindele_driverbehavior_ITS2015,Alahi_sociallstm_cvpr2016,Schulz_interaction_DM_IV2019,Chandra_Traphic_CVPR2019}. 
%
%
%
%
However, most of the existing work still cannot effectively model complex interactions since many of them are leveraging  ``hand-crafted interaction models''~\cite{Alahi_sociallstm_cvpr2016}. 
%
%
Data-driven approaches are better options as they can learn subtle and complex interactions~\cite{Alahi_sociallstm_cvpr2016,Vemula_socialattention_icra2018,Chen_crowdnavi_icra2019,Chandra_Traphic_CVPR2019}. However, existing approaches are still insufficient for three reasons. 

First, the input used by several existing methods~\cite{Alahi_sociallstm_cvpr2016,Vemula_socialattention_icra2018,Chen_crowdnavi_icra2019} is the human's 2D-location on bird's-eye-view (BEV) images. 
However, it is more desirable to use ego-perspective sensing devices, e.g, cameras, as humans use two eyes to sense.
This calls for a specific design for egocentric interaction models.
Second, using 2D pixel coordinates to model the 3D interactions (such as \cite{Chandra_Traphic_CVPR2019}) is insufficient because of perspective projection. 
BEV images can resolve this problem since the depth and spatial positions are both embedded in the BEV images.
%
%
Third, the existing approaches only consider human-human or human-robot interactions, ignoring the environment factors, such as lane markings, crosswalks, and traffic lights. 
However, modeling these objects is nontrivial because they have irregular shapes.
%
%

%

%
In this paper, we propose a 3D-aware egocentric spatial-temporal interaction framework for automated driving applications.
Our method is the first method based on egocentric images and can address the aforementioned problems. 
The specific approach we take is to design two graph convolutional networks (GCN)~\cite{Kipf_GCN_iclr2017} to model the egocentric interactions.
We define two graphs, \textit{Ego-Thing Graph} and \textit{Ego-Stuff Graph} to encode how the ego vehicle interacts with the thing objects (e.g., cars and pedestrians) and the stuff objects (e.g., lane markings and traffic lights).
%
%
The ego-thing graph is an improvement of Wu et al. ~\cite{Wu_groupGCN_cvpr2019}.
We introduce two new concepts. 
We add an ego node (i.e., the ego vehicle) for egocentric interaction modeling, and we incorporate the objects' 3D locations (recovered from image-based depth estimation). 
The ego-stuff graph is designed similarly. 
However, in order to extract features from irregular stuff objects, we introduce a new method known as the MaskAlign operation.
%
%
%


%

We validate the proposed framework on tactical driver behavior recognition using Honda Research Institute Driving Dataset (HDD)~\cite{Ramanishka_behavior_CVPR_2018}. 
The HDD is the largest dataset in the field. 
It provides 104-hour egocentric videos with frame-level annotations of tactical driver behavior.
We validate our method based on two types of settings: 1) the ego vehicle has interactions with stuff objects (e.g., lane change, lane branch, and merge) and 2) the ego vehicle has interactions with thing objects (e.g., stop for crossing pedestrian and deviate for parked car).
Our approach offers substantial performance boost (in terms of mAP, See Experiment section for definitions) over baselines on the two settings by 3.9\% and 6.0\%, respectively. 
%
%
%

\section{RELATED WORKS}




\subsection{Tactical Driver Behavior Recognition}
%
Significant efforts have been made in tactical driver behavior recognition~\cite{OliverIV2000,Kuge2000ADB,Mitrovic_ITS_2005,Wu_ITS_2013,Jain_anti_maneuver_iccv2015,XuCVPR2017,wang2018deep,Ramanishka_behavior_CVPR_2018,Xu_TRN_iccv2019}. 
Hidden Markov networks (HMM) were leveraged to recognize driver behaviors~\cite{OliverIV2000,Kuge2000ADB,Mitrovic_ITS_2005,Wu_ITS_2013,Jain_anti_maneuver_iccv2015}. 
A single node in HMM encodes the states from the ego vehicle, roads and traffic participants~\cite{OliverIV2000} into a state vector. 
In the proposed framework, we explicitly model the above three states using different nodes, each of which encodes its own representation according to the semantic context. 
Recently, convolutional and recurrent neural network based algorithms~\cite{XuCVPR2017,Ramanishka_behavior_CVPR_2018,Xu_TRN_iccv2019} are proposed. 
They implicitly encode the states of the ego vehicle and road users using 2D convolution, and the state transition is via recurrent units. 
Our method explicitly models the states using graph convolutional networks (GCN) and uses temporal convolution networks for the state transition.

Wang et al.,~\cite{wang2018deep} designed an object-level attention layer to capture the impacts of objects on driving policies. 
Instead of simply weighting and concatenating objects' features, our framework preserves more complicated forms of interactions benefiting from GCN.
Additionally, interactions between the ego vehicle and road infrastructure are included in our system.



\subsection{Graph Neural Networks for Driving Scenes}
Recently, graph neural networks (GNN)~\cite{Li_GGNN_iclr2016,Kipf_GCN_iclr2017} has made significant progress in situation recognition~\cite{Li_situatino_iccv2017}, action recognition~\cite{Yan_stgcn_aaai2018,Wang_videospacetime_eccv2018}, group activity recognition~\cite{Wu_groupGCN_cvpr2019}, and scene graph generation~\cite{Yang_aGCN_eccv2018}. 
However, considerably less attention has been paid to driving scene applications 
%

%
Herzig et al.~\cite{Herzig_collision_arxiv2018} proposed a Spatio-Temporal Action Graph (STAG) network to detect driving collision. 
While STAG is similar to the proposed \textit{Ego-thing graph}, our model explicitly exploits 3D locations of objects and the ego vehicle into the design of nodes and edges. 
The 3D cue is essential in understanding scenes from egocentric perspective. 
This design is motivated by~\cite{Wu_groupGCN_cvpr2019}.
Note that 2D locations are used in~\cite{Wu_groupGCN_cvpr2019} while we use 3D locations extracted from~\cite{Lasinger_mix_depth_2019}.
Moreover, we consider interactions between the ego vehicle and road infrastructure that enable the proposed framework to be applied for diverse driving scene applications, e.g., learning driving model from images~\cite{Matthias_Modularity_corl2018}.
The details of our graph design can be found in Section~\ref{subsection:ego-thing graph}.

\section{Egocentric Spatial-Temporal Interaction Modeling}
\begin{figure*}[h!]
    \centering
    \includegraphics[width=0.9\textwidth]{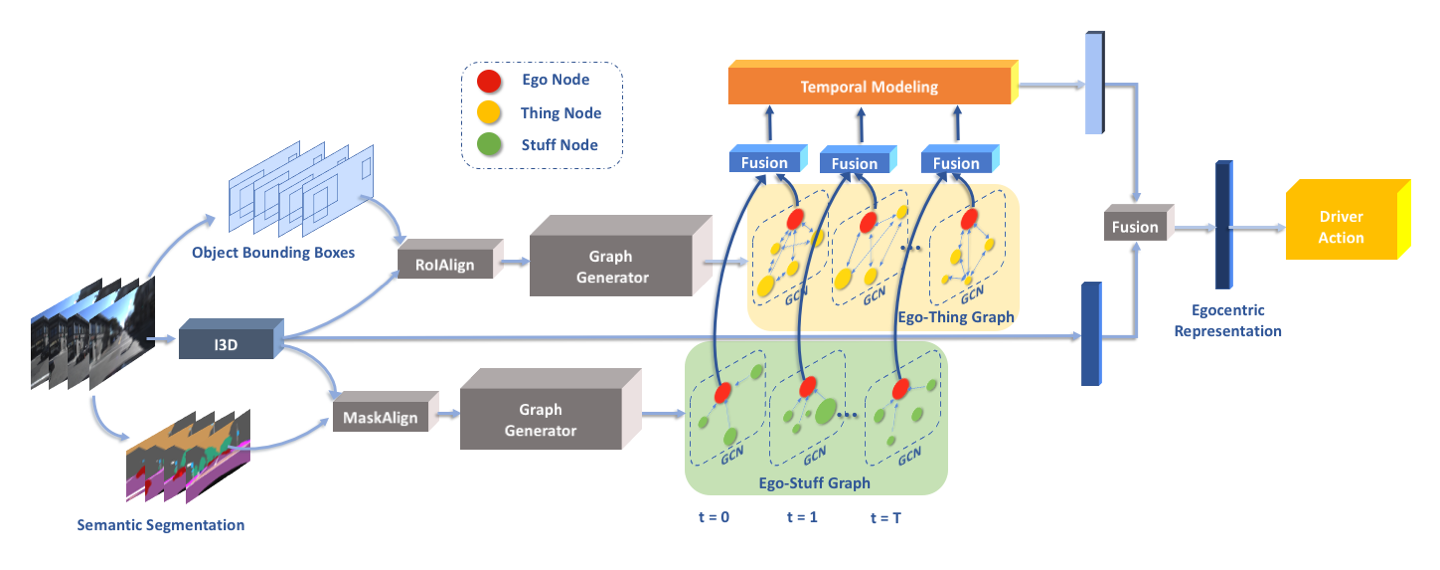}
    \caption{An overview of our framework. Given a video segment, our model applies 3D convolutions to extract visual features followed by two branches: RoIAlign is employed to extract object features from object bounding boxes and MaskAlign is designed to extract features of irregular shape objects from semantic masks. Then, frame-wise \textit{Ego-Thing Graph} and \textit{Ego-Stuff Graph} are constructed to propagate interactive information among objects via graph convolution networks. The outputs of the two graphs are fused and fed into a temporal fusion module to form interactive representation. Finally, global video representation from I3D head and interactive features are aggregated as an input to tactical driver behavior recognizer.}
    \label{fig:pipeline}
    \vspace{-10pt}
\end{figure*}

\subsection{Overall Architecture}
%
An overview of the proposed framework is depicted in Fig.~\ref{fig:pipeline}.
Given video frames, we apply instance segmentation and semantic segmentation in \cite{HeCVPR2017} to obtain \textit{thing objects} and \textit{stuff objects}, respectively.
%
Object features are extracted from intermediate I3D \cite{CarreiraCVPR2017} features via RoIAlign \cite{HeCVPR2017} and MaskAlign (Section \ref{subsection:ego-stuff graph}).
Afterwards, we construct \textit{Ego-Thing Graphs} and \textit{Ego-Stuff Graphs} in a timely manner and apply graph convolutional networks (GCN) \cite{Kipf_GCN_iclr2017} for message passing.
The updated ego features from two graphs are fused and processed via a temporal fusion module.
Additionally, the temporally fused ego features are concatenated with the I3D head feature, which serves as a global video embedding, to form the egocentric representation. 
At last, this egocentric feature is passed through a fully connected layer to obtain the final classification.

\subsection{Ego-Thing Graph}
\label{subsection:ego-thing graph}
%
The ego-thing graph is designed to model interactions among ego vehicle and movable traffic participants, such as $\langle \textit{car, ego vehicle} \rangle,\langle \textit{car, person} \rangle$ and so on.

\textbf{Node feature extraction.} 
In our design, \textit{thing objects} are \textit{car}, \textit{person}, \textit{bicycle}, \textit{motorcycle}, \textit{bus}, \textit{train}, and \textit{truck}. 
Given bounding boxes generated from Mask R-CNN \cite{HeCVPR2017}, we keep the top K detections on each frame from all the classes above and set K to 20.
%
%
Then RoIAlign \cite{HeCVPR2017} and a max pooling layer are applied to obtain $1 \times D$ dimensional appearance features as \textit{Thing Node} features in a ego-thing graph. 
The \textit{Ego Node} feature is obtained by the same procedure from a frame-size bounding box. 

\textbf{Graph definition.} 
We denote the sequence of frame-wise ego-thing graphs as $\mathbf{G}^{ET} = \{\mathbf{G}^{ET}_t | t =1,\cdots,T\}$, where $T$ is the number of frames, and $\mathbf{G}^{ET}_t\in \mathbb{R}^{(K+1) \times (K+1) }$ is the ego-thing affinity matrix at frame $t$ representing the pair-wise interactions among \textit{thing objects} and ego.
Specifically, ${G}^{ET}_t(i,j)$ denotes the influence of object $j$ on object $i$.  
Nodes in graph correspond to a set of objects $\{ (\mathbf{x}^{t}_i,\mathbf{p}^t_i) | i = 1, \cdots, K+1\}$, where $\mathbf{x}^{t}_i \in \mathbb{R}^{D} $ is $i$-th object's appearance feature, and $\mathbf{p}^t_i \in \mathbb{R}^{3}$ is the 3D location of the object in world frame.
Note that index $K+1$ corresponds to ego object and $i=1,\cdots,K$ correspond to \textit{thing objects}.

\textbf{Interaction modeling.} 
Ego-thing interactions are defined as second-order interactions, where not only the original state but also the changing state of the \textit{thing object} caused by other objects will altogether influence the ego state.
To sufficiently model these interactions, we consider both appearance features and distance constraints inspired by \cite{Wu_groupGCN_cvpr2019}. 
We compute the edge value $G_t^{ET}(i,j)$ as:
\begin{equation}
    {G}^{ET}_t(i,j) = \frac{f_s(\mathbf{p}^t_i,\mathbf{p}^t_j)\text{exp}(f_a(\mathbf{x}^{t}_i,\mathbf{x}^{t}_j))}{\sum_{j=1}^{K+1}  f_s(\mathbf{p}^t_i,\mathbf{p}^t_j)\text{exp}(f_a(\mathbf{x}^{t}_i,\mathbf{x}^{t}_j))} 
    \label{eq:1}
\end{equation}
where $f_a(\mathbf{x}^{t}_i,\mathbf{x}^{t}_j)$ indicates the appearance relation between two objects, and we set up a distance constraint via a spatial relation  $f_s(\mathbf{p}^t_i,\mathbf{p}^t_j)$. 
Softmax function is used to normalize the influence on object $i$ from other objects.

The appearance relation is calculated as below:
\begin{equation}
    f_a(\mathbf{x}^{t}_i,\mathbf{x}^{t}_j) = \frac{\phi(\mathbf{x}^{t}_i)^{\text{T}} \phi '(\mathbf{x}^{t}_j)}{\sqrt{D}}
\end{equation}
where $\phi(\mathbf{x}^{t}_i) =\mathbf{w}\mathbf{x}^{t}_i $ and $\phi '(\mathbf{x}^{t}_j) = \mathbf{w}'\mathbf{x}^{t}_j$. 
Both $\mathbf{w} \in \mathbb{R}^{D \times D}$ and $\mathbf{w}'\in \mathbb{R}^{D \times D}$ are learnable parameters which map appearance features to a subspace and enable learning the correlation of two objects. 
$\sqrt{D}$ is a normalization factor.

 The necessity of defining spatial relation arises from that the interactions of two distant objects are usually scarce. 
 %
 %
To calculate this relation, we first unproject objects from 2D image plane to the 3D space in the world frame \cite{Lasinger_mix_depth_2019}:
\begin{equation}
\begin{bmatrix} x & y & z & 1 \end{bmatrix}^T  =\delta_{u,v} \cdot \mathbf{P}^{-1}\begin{bmatrix} u & v & 1 \end{bmatrix}^T 
\end{equation}
where $\begin{bmatrix} u & v & 1 \end{bmatrix}^{T}$ and $\begin{bmatrix} x & y & z & 1 \end{bmatrix}^{T}$ are homogeneous representations in 2D and 3D coordinate systems, $\mathbf{P}$ is the camera intrinsic matrix, and $\delta_{u,v}$ is the relative depth at $(u,v)$ obtained by depth estimation \cite{LasingerArxiv2017}.
In the 2D plane, we choose the centers of bounding boxes to locate \textit{thing objects}. 
The location of the ego vehicle is fixed at the middle-bottom pixel of the frame.
Then the spatial relation function $f_s$ is formulated as:
 \begin{equation}
    f_s(\mathbf{p}^t_i,\mathbf{p}^t_j) = \mathbb{I}(d(\mathbf{p}^t_i,\mathbf{p}^t_j)\leq \mu)
    \label{eq:4}
\end{equation}
where $\mathbb{I}(\cdot)$ is the indicator function, $d(\mathbf{p}^t_i,\mathbf{p}^t_j)$ computes the Euclidean distance between object $i$ and object $j$ in the 3D space, and $\mu$ is the distance threshold which regulates the spatial relation value to be zero if the distance is beyond this upper bound. 
In our implementation, the value of $\mu$ is set to be 3.0.
\subsection{Ego-Stuff Graph}  \label{subsection:ego-stuff graph}
%
The ego-stuff graph $\mathbf{G}^{ES}$ is constructed in a similar manner as the ego-thing graph $\mathbf{G}^{ET}$  in Eq.\ref{eq:1} except for the following aspects:

\textbf{Node feature extraction.} 
We include the following classes as \textit{stuff objects}: \textit{Crosswalk}, \textit{Lane Markings}, \textit{Lane Separator}, \textit{Road}, \textit{Service Lane}, \textit{Traffic Island}, \textit{Traffic Light} and \textit{Traffic Sign}. 
The criterion we use to distinguish \textit{stuff objects} from \textit{thing objects} is based on whether the change of states can be caused by other objects. 
For example, cars stop and yield to person, but a traffic light turns red to green by itself.
Another distinction lies in that the contour of most \textit{stuff objects} cannot be well depicted as rectangular bounding boxes. 
%
%
Thus, it is difficult either to detect it by algorithms like Faster R-CNN \cite{RenNeurIPS2015}, YOLO \cite{RedmonCVPR2016} or to extract features by RoIAlign \cite{HeCVPR2017} without enclosing irrelevant information.
For this, we propose a feature extraction approach named MaskAlign to extract features for a binary mask $\mathbf{M}_i^{t}$, which is the $i$-th \textit{stuff object} at time $t$. 
$\mathbf{M}_i^{t}$ is downsampled to ${\mathbf{M}_i^{t}}'$ ($W\times H$) with the same spatial dimension as the intermediate I3D feature map $\mathbf{X}$  $(T \times W \times H \times D)$.
We compute the \textit{stuff object} feature by MaskAlign as following:
 \begin{equation}
    \mathbf{x}_i^{t} = \frac{\sum_{w=1}^{W}\sum_{h=1}^{H} \mathbf{X}^{t}_{(w,h)} \cdot {{\mathbf{M}_{i}^{t}}'}_{(w,h)} }{\sum_{w=1}^{W}\sum_{h=1}^{H} {{\mathbf{M}_{i}^{t}}'}_{(w,h)}}
\end{equation}
where $\mathbf{X}^{t}_{(w,h)} \in \mathbb{R}^{1\times D}$ is the D-dimension feature at pixel $(w,h)$ for time $t$, and ${{\mathbf{M}_{i}^{t}}'}_{(w,h)}$ is a binary scalar indicating whether object $i$ exists at pixel $(w,h)$.

\textbf{Interaction Modeling.} In ego-stuff graph, we ignore interactions among \textit{stuff objects} since they are insusceptible to other objects.
Hence, we set $f_s$ to zeros for every pair of \textit{stuff objects} and only pay attention to the influence that \textit{stuff objects} act on ego vehicle.
We call it as the first-order interaction.
To better model the spatial relations, instead of unprojecting bounding box centers, we map every pixel inside the downsampled binary mask ${{\mathbf{M}_{i}^{t}}'}$ to 3D space and calculate the Euclidean distance between every pixel with the ego vehicle. 
The distance is the minimum distance of the all. 
The distance threshold in ego-stuff graph is designed as 0.8.

\subsection{Reasoning on Graphs}
To perform reasoning on graphs, we introduce graph convolutional networks (GCN) proposed in \cite{Kipf_GCN_iclr2017}. 
GCN takes a graph as input, passes information through the learned edges, and refreshes nodes' features as output. 
Specifically, graph convolution can be expressed as:

 \begin{equation}
    \mathbf{Z}^{l+1} =  \mathbf{G}\mathbf{Z}^{l}\mathbf{W}^{l}+\mathbf{Z}^{l}
\end{equation}
where $\mathbf{G}$ is the affinity matrix from graphs.
Taking ego-thing graph as an example, $\mathbf{Z}^{l} \in \mathbb{R}^{(K+1)\times D}$ is the appearance feature matrix of nodes in the $l$-th layer.
$\mathbf{W}^{l} \in \mathbb{R}^{ D \times D}$ is the learnable weight matrix. 
We also build a residual connection by adding $\mathbf{Z}^{l}$. 
%
%
In the end of each layer, we adopt Layer Normalization \cite{BaArxiv2016} and ReLU before  $\mathbf{Z}^{l+1}$ is fed to the next layer.
As second-order interaction is not considered in ego-stuff graph but in ego-thing graph, we use one layer GCN in ego-stuff graph and two layers in ego-thing graph. 

\subsection{Temporal Modeling}
\begin{figure}
    \centering
    \includegraphics[width=\columnwidth]{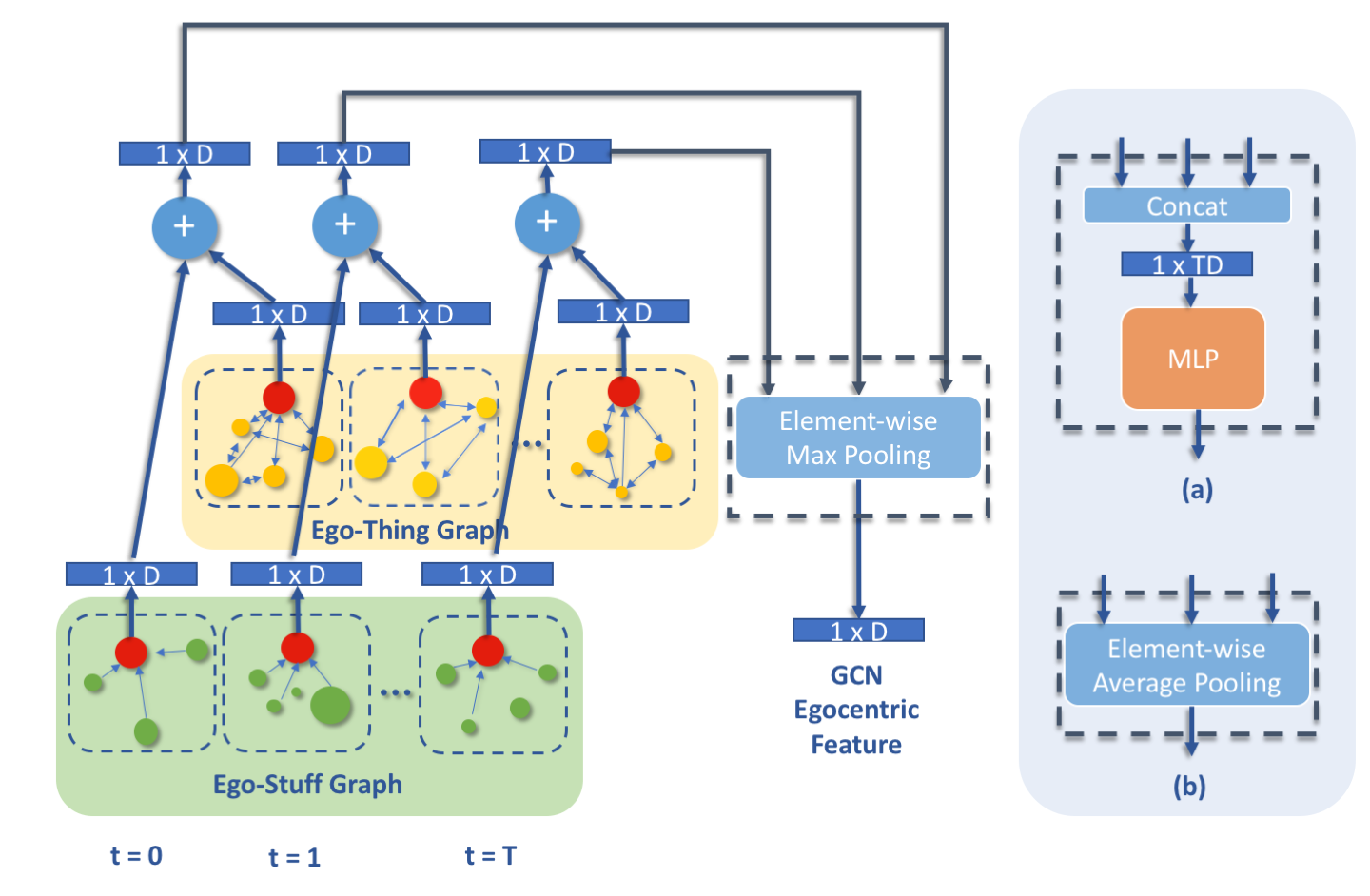}
    \vspace{-16pt}
    \caption{Architecture of our temporal modeling module. }
    \label{fig:temporalmodeling}
    \vspace{-6pt}
\end{figure}
GCN interactive features in each frame are processed independently without considering temporal context information. 
Therefore, we append a temporal fusion module to the late stage in our framework as illustrated in Fig.~\ref{fig:temporalmodeling}.
Unlike prior works \cite{Wang_videospacetime_eccv2018, Wu_groupGCN_cvpr2019,Yan_stgcn_aaai2018}, which fuse features of every node in different graphs, we only focus on ego node. 
Ego features are aggregated by a element-wise summation from two types of graphs. 
Then these time-specific ego features are fed into a temporal fusion module, which applies element-wise max pooling to obtain a $1\times D$ feature vector, namely GCN egocentric feature.
We also propose another two designs for temporal fusion:
(a) Inspired by Temporal Relation Network \cite{ZhouECCV2018}, which utilizes multi-layer perceptrons (MLP) as temporal modeling, we follow the similar approach in order to capture the temporal ordering of patterns.   
(b) The temporal fusion can also be replaced by element-wise average pooling. 
In Section \ref{subsection:ablation_studies}, we conduct different experiments to investigate all three temporal modeling approaches.

%
%
%
%

\section{EXPERIMENTS}
\subsection{Dataset}
We evaluate the proposed framework on the HDD dataset \cite{Ramanishka_behavior_CVPR_2018}, the largest dataset that provides 104-hour egocentric videos with frame-level annotations of tactical driver behavior. It has a diverse set of scenarios where complicated interactions happen between the ego vehicle and road users.
%
%
The data was collected within San Francisco Bay Area including urban, suburban and highways. 
We follow the same Train/Test data split as \cite{Ramanishka_behavior_CVPR_2018}.
%

The videos are labeled by a 4-layer representation to describe tactical driver behaviors. 
Among these 4 layers, \textbf{Goal-oriented action} layer (e.g., left turn and right lane lane change) and \textbf{Cause} layer (e.g., stop for crossing vehicle) consist of the actions with interactions.
We leverage those labels and analyze the effectiveness of the proposed interaction modeling framework in Section \ref{subsection:analysis_on_interactions}.
%

%
%
%
%
%
%
%

\begin{table*}[t]
    \small
    \centering
    \vspace{5pt}
    \resizebox{\textwidth}{!}{
        
        \begin{tabular}
            {@{}lc@{\;}c@{\;}@{\;}c@{\;}@{\;}c@{\;}@{\;}c@{\;}@{\;}c@{\;}@{\;}c@{\;}@{\;}c@{\;}@{\;}c@{\;}@{\;}c@{\;}@{\;}c@{\;}@{\;}c@{\;}@{\quad\;\;\;\;}c@{}}
            
\toprule
& & \multicolumn{11}{c}{Individual actions} & \\
\cmidrule(r{4\cmidrulekern}){3-13}
            \begin{tabular}{@{}c@{}} \\ Method \end{tabular} &
            \begin{tabular}{@{}c@{}} \\ Online/Offline \end{tabular} &
            \begin{tabular}{@{}c@{}}intersection \\ passing \end{tabular} &
            \begin{tabular}{@{}c@{}}\\ L  turn            \end{tabular} &
            \begin{tabular}{@{}c@{}}\\ R  turn           \end{tabular} &
            \begin{tabular}{@{}c@{}}L  lane \\ change  \end{tabular} &
            \begin{tabular}{@{}c@{}} R  lane \\ change \end{tabular} &
            \begin{tabular}{@{}c@{}} L  lane \\ branch  \end{tabular} &
            \begin{tabular}{@{}c@{}} R  lane \\ branch \end{tabular} &
            \begin{tabular}{@{}c@{}}crosswalk \\ passing    \end{tabular} &
            \begin{tabular}{@{}c@{}}railroad \\ passing     \end{tabular} &
            \begin{tabular}{@{}c@{}} \\merge                   \end{tabular} &
            \begin{tabular}{@{}c@{}} \\u-turn                  \end{tabular} &
            \begin{tabular}{@{}c@{}}Overall \\mAP                  \end{tabular} \\
            \midrule
            \cmidrule{1-14}
            CNN~\cite{Ramanishka_behavior_CVPR_2018} & \multirow{5}{*}{Online} & 53.4 & 47.3 & 39.4 & 23.8 & 17.9 & 25.2 & 2.9 & 4.8 & 1.6 & 4.3 & 7.2 & 20.7 \\
            CNN-LSTM~\cite{Ramanishka_behavior_CVPR_2018} & & 65.7 & 57.7 & 54.4 & 27.8 & 26.1 & 25.7 & 1.7 & 16.0 & 2.5 & 4.8 & 13.6 & 26.9 \\
            ED~\cite{Xu_TRN_iccv2019} & & 63.1 & 54.2 & 55.1 &28.3 & 35.9 & 27.6 & 8.5 & 7.1 & 0.3 & 4.2 & 14.6 & 27.2 \\
            TRN~\cite{Xu_TRN_iccv2019} & & 63.5 & 57.0 & 57.3 & 28.4 & 37.8 & 31.8 & 10.5 & 11.0 & 0.5 & 3.5 & 25.4 & 29.7 \\
            DEPSEG-LSTM~\cite{Narayanan_tatical_cvprw2018} & & 70.9 & 63.4 & 63.6 & 48.0 & 40.9 & 39.7 & 4.4 & 16.1 & 0.5 & 6.3 & 16.7 & 33.7 \\
            C3D~\cite{Tran_C3D_iccv2015} & & 72.8&	64.8	&71.7&	53.4	&44.7	&52.2	&3.1	&14.6	&2.9	&10.6	&15.8	&37.0 \\
            \cmidrule{1-14}
            \cmidrule{1-14}
                \cline{6-9}\cline{12-12}
            C3D~\cite{Tran_C3D_iccv2015} & \multirow{3}{*}{Offline} & 82.4	& 77.4 & \textbf{80.7} &\multicolumn{1}{|c}{ 67.9}	& 56.9	&59.7	& \multicolumn{1}{c|}{5.2}	&17.4	&\textbf{3.9} &	\multicolumn{1}{|c|}{20.1}	&29.5&	45.5 \\
         
            I3D~\cite{CarreiraCVPR2017} & & \textbf{85.6} & \textbf{79.1} & 78.9 &  \multicolumn{1}{|c}{74.0} & \textbf{62.4} & 59.0 & \multicolumn{1}{c|}{14.3} & \textbf{29.8} & 0.1 & \multicolumn{1}{|c|}{20.1} & \textbf{41.4} & 49.5 \\
            Ours & & 85.5 & 77.9 & 79.1 & \multicolumn{1}{|c}{\textbf{76.0}} & 62.0 & \textbf{64.0} &\multicolumn{1}{c|}{ \textbf{19.8}} & 29.6 & 1.0 & \multicolumn{1}{|c|}{\textbf{27.7}} & 39.0 & \textbf{51.1} \\
             \cline{6-9}\cline{12-12}
\bottomrule
        \end{tabular}
    }
    \caption{\textit{Results of \textbf{\textrm{Goal-oriented}} driver behavior recognition on HDD.} The unit is \%.}
    \vspace{-15pt}
    \label{table:hdd_detection}
\end{table*}

\subsection{Implementation Details}
We implemented our framework in TensorFlow. All experiments are performed on a server with 4 NVIDIA TITAN-XP.
%
%
The input to the framework is a 20-frame clip with a resolution of $224 \times 224$ at 3 fps, approximately 6.67s.
We adopt Inception-v3 \cite{Szegedy_inceptionv3_cvpr2016} pre-trained on ImageNet~\cite{Russakovsky_imagenet_ijcv2015} as the backbone, following \cite{CarreiraCVPR2017} to inflate 2D convolution into a 3D ConvNet, and fine-tune it on the Kinetics action recognition dataset \cite{Kay_kinetics_arvix2017}. 
The intermediate feature map used in RoIAlign and MaskAlign is extracted from the \texttt{Mixed\_3c} layer, where $D = 512$ is the number of feature channels. 
The global I3D feature is generated from a $1 \times 1 \times 1$ convolution on \texttt{Mixed\_5c} layer feature, which reduces the output channel number from 1024 to 512. 
The downsampled binary mask ${{\mathbf{M}_{i}^{t}}'}$ is $28\times 28$.
The model is trained in a two-stage training scheme with batch size set to 32: (1) we fine-tune the Kinetics pre-trained model on the HDD dataset for 50K iterations without using GCN. 
We refer to this model the baseline for our experiment. 
(2) We load the weights trained in Stage 1, and further train the network together with GCN for 20K iterations. 
 \begin{table}[t]
    \small
    \centering
    \resizebox{\columnwidth}{!}{
        \begin{tabular}
            {@{}lc@{\;}@{\;}c@{\;}@{\;}c@{\;}@{\;}c@{\;}@{\;}c@{\;}@{\;}c@{\;}@{\quad\;\;\;\;}c@{}}
\toprule
& \multicolumn{6}{c}{Individual actions} & \\
\cmidrule(r{4\cmidrulekern}){2-7}
            \begin{tabular}{@{}c@{}@{}} \\ Method \end{tabular} &
            \begin{tabular}{@{}c@{}@{}}  \\ Stop for \\ Congestion  \end{tabular} &
            \begin{tabular}{@{}c@{}@{}} \\Stop for \\Sign           \end{tabular} &
            \begin{tabular}{@{}c@{}} \\Stop for \\Red Light           \end{tabular} &
            \begin{tabular}{@{}c@{}@{}} Stop for \\ Crossing \\ Vehicle  \end{tabular} &
            \begin{tabular}{@{}c@{}@{}}Deviate for \\ Parked \\ Vehicle \end{tabular} &
            \begin{tabular}{@{}c@{}@{}} Stop for \\Crossing  \\Pedestrian \end{tabular} &
            \begin{tabular}{@{}c@{}}Overall \\mAP                  \end{tabular} \\
            \midrule

            I3D~\cite{CarreiraCVPR2017} & 64.8 & 71.7 & 63.6 &21.5& 15.8 & 26.2  & 43.9 \\
            Ours & \textbf{74.1} & \textbf{72.4} & \textbf{76.3} & \textbf{26.9} & \textbf{20.4} & \textbf{29.0} & \textbf{49.9} \\

\bottomrule
        \end{tabular}
    }
    \caption{\textit{Results of driver behavior recognition in \textbf{\textrm{Cause}} layer on HDD.} The unit is \%.}
    \vspace{-5pt}
    \label{table:hdd_detection_cause}
\end{table}
We use Adam \cite{Kingma_adam_arvix2014} optimizer with default parameters. We set learning rate as 0.001 and 0.0002 for the first and second stage for training, respectively.

\subsection{Analysis on Interactions}\label{subsection:analysis_on_interactions}
To understand the benefits of modeling interactions, we perform analysis on the following two aspects.

\textbf{Goal-oriented Action Layer.} 
Table~\ref{table:hdd_detection} presents \textbf{Goal-oriented} action recognition results. 
We use the per-frame mean average precision (mAP) as evaluation metric in all experiments. 
We pay attention to the 5 `lane-related' classes in frames: \textit{Left Lane Change}, \textit{Right Lane Change}, \textit{Left Lane Branch}, \textit{Right Lane Branch} and \textit{Merge}. 
 Our model obtains 49.9\% mAP over these 5 classes, which surpasses the I3D baseline 46.0\% mAP by a gain of 3.9\%. 
 This improvement showcases the effectiveness of modeling interactions between ego vehicle and traffic lanes, which also can be validated by visualization in Section \ref{subsection:visualization}.

\begin{figure}[h!]
\vspace{5pt}
\minipage{\columnwidth}
  \includegraphics[width=\linewidth,trim={0 0.3cm 0 2cm },clip]{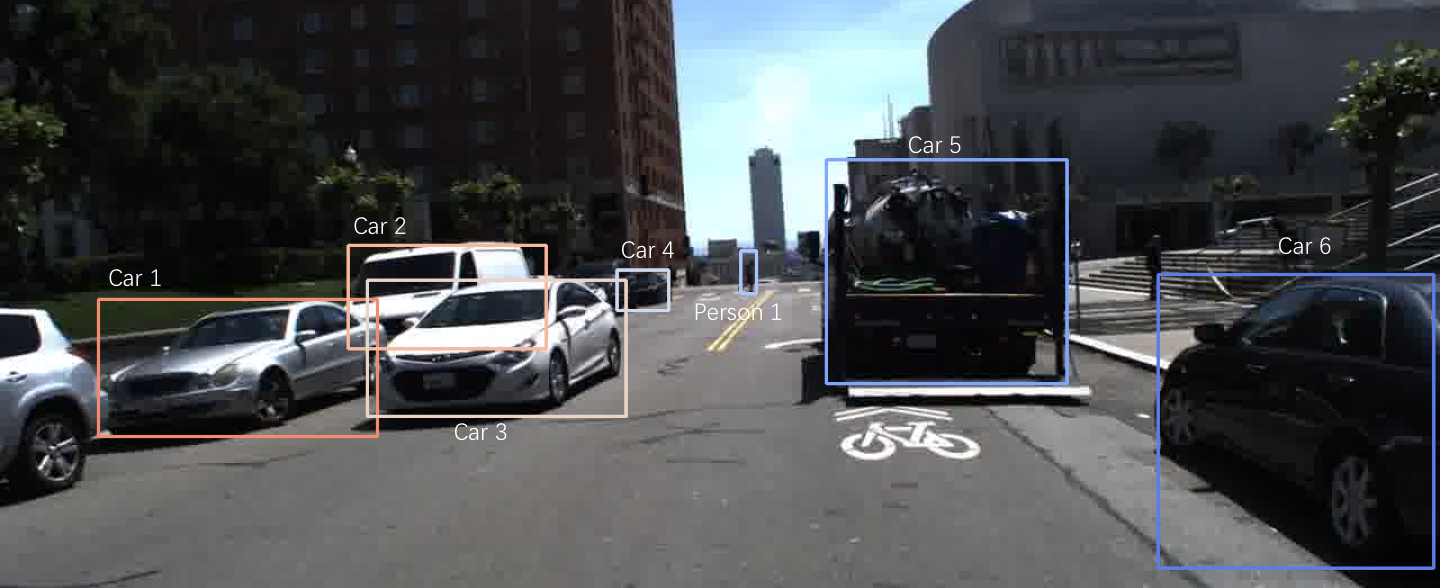}
  \vspace{-15pt}
  \caption*{ \footnotesize (a) \textbf{Goal-oriented} prediction: Background and \textbf{Cause} prediction: Parked Vehicle}
\endminipage \hfill

\minipage{0.33\columnwidth}
\vspace{0pt}
  \includegraphics[width=\linewidth]{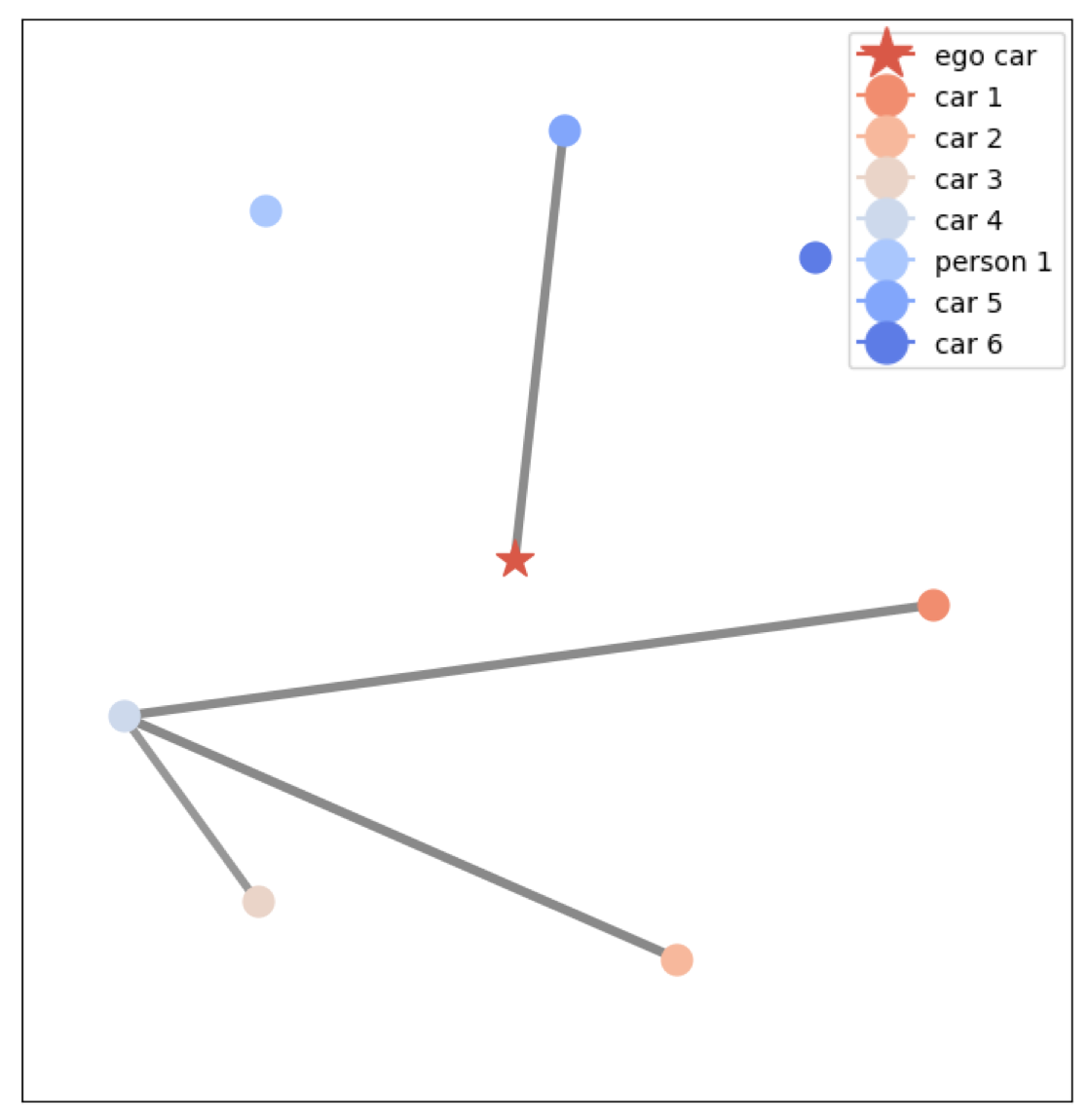}
  \caption*{\footnotesize(b) Learned affinity \\ matrix}
\endminipage\hfill
\minipage{0.66\columnwidth}
 \vspace{0pt}
  \includegraphics[height=0.5\linewidth]{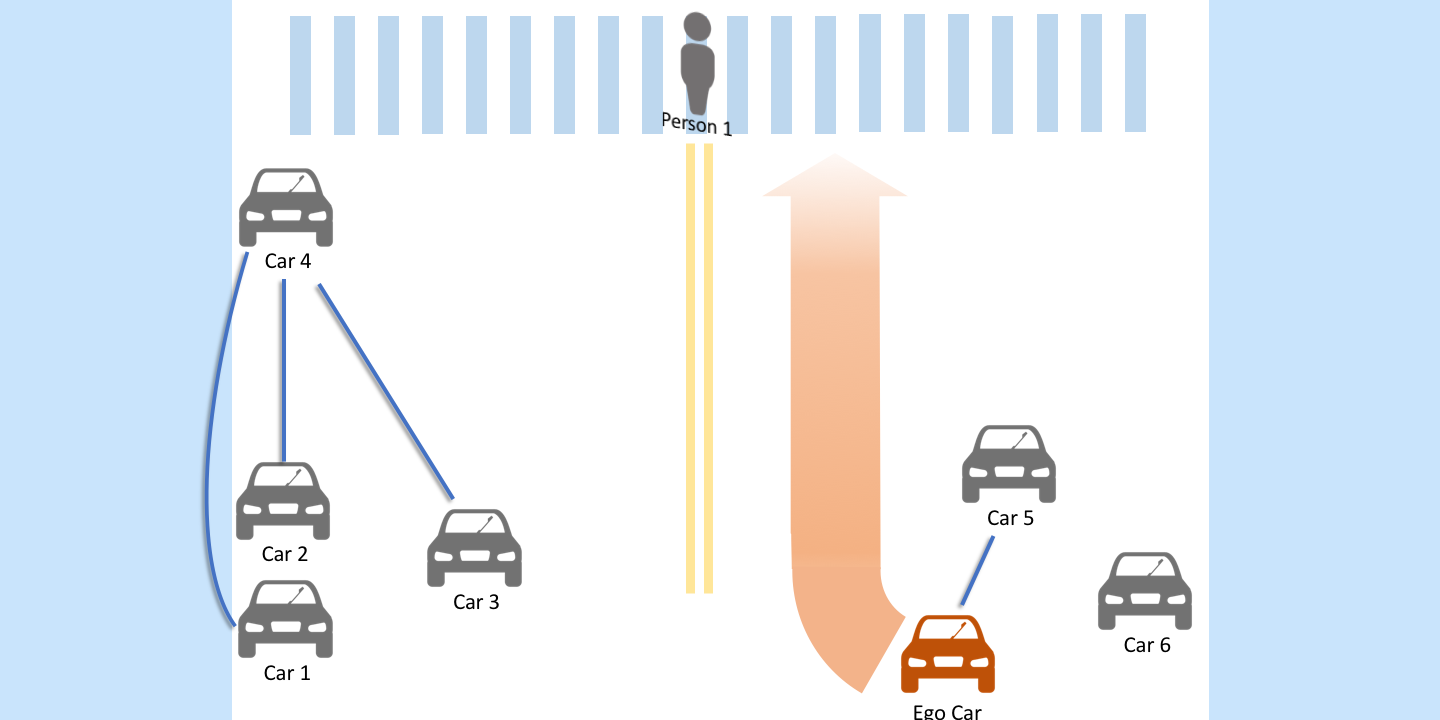}
  \caption*{\footnotesize(c) Visualization of the learned affinity matrix using a top-view scene representation}
\endminipage
 \vspace{-4pt}
 \caption{Attention visualization from top-view.}
 \label{fig:visualization_scene_representation2}
 \vspace{-12pt}
\end{figure}

 \textbf{Cause Layer.} 
6 classes from \textbf{Cause} layer are designed to explain the reason for \textit{stop} and \textit{deviate} actions, such as \textit{Deviate for Parked Vehicle}, which is an example of ego-thing interaction.
We extend our framework to multi-head classifiers to simultaneously predict \textbf{Goal-oriented} actions and \textbf{Causes}.
Note that we train a multi-head I3D as the baseline for this experiment.
Our design achieves a steady increase in recognizing \textbf{Goal-oriented} actions by improving the baseline of 48.5\% to 50.2\%.
 Meanwhile, the result of \textbf{Cause} layer in Table~\ref{table:hdd_detection_cause} shows a significant gain of 6.0\% in overall mAP.
 We further demonstrate the strength of the proposed interaction modeling by using a \textit{Deviate for Parked Vehicle} scenario in Fig.~\ref{fig:visualization_scene_representation2} in Section \ref{subsection:visualization}.
 
\subsection{Comparison with the State of the Art}
 We compare our approach with the state-of-the-art in Table \ref{table:hdd_detection}.
  We categorize the existing methods tested on HDD into \textit{online} and \textit{offline}.
The online approaches aim to detect driver actions as soon as a frame arrives.
Future context is not considered.
%
The offline approaches take future frames into consideration.
 Since future information is processed, the offline approaches exhibit an overwhelming advantage over the online approaches.
 Among the offline methods, our model significantly outperforms the C3D \cite{Tran_C3D_iccv2015} and I3D \cite{CarreiraCVPR2017} by 5.6\%  and 1.6\% in terms of mAP, respectively. 

\begin{figure*}[h!]
\vspace{3pt}
\minipage{0.25\textwidth}
  \includegraphics[width=\linewidth,trim = {0 0.8cm 0 0.8cm},clip]{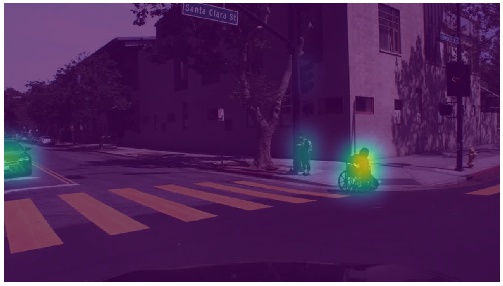}
  \caption*{(a) Left Turn}
\endminipage\hfill
\minipage{0.25\textwidth}
  \includegraphics[width=\linewidth,trim = {0 0.8cm 0 0.8cm},clip]{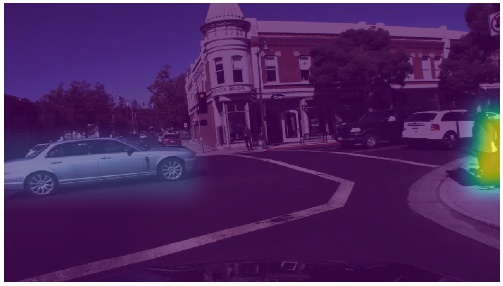}
  \caption*{(b) Right Turn}
\endminipage\hfill
\minipage{0.25\textwidth}
  \includegraphics[width=\linewidth,trim = {0 0.8cm 0 0.8cm},clip]{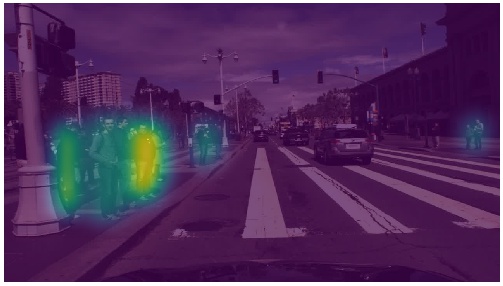}
  \caption*{(c) Crosswalk Passing}
\endminipage\hfill
\minipage{0.25\textwidth}%
  \includegraphics[width=\linewidth,trim = {0 0.8cm 0 0.8cm},clip]{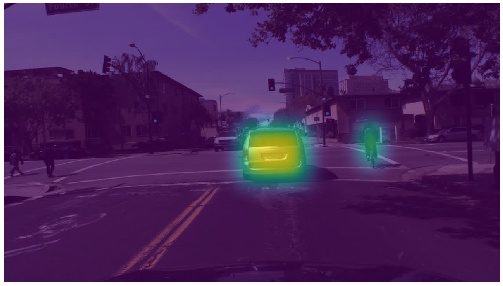}
  \caption*{(d) Intersection Passing}
\endminipage \hfill

\minipage{0.25\textwidth}
  \includegraphics[width=\linewidth,trim = {0 0.8cm 0 0.8cm},clip]{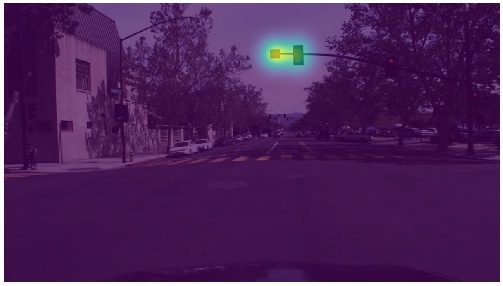}
  \caption*{(e) Left Turn}
\endminipage\hfill
\minipage{0.25\textwidth}
  \includegraphics[width=\linewidth,trim = {0 0.8cm 0 0.8cm},clip]{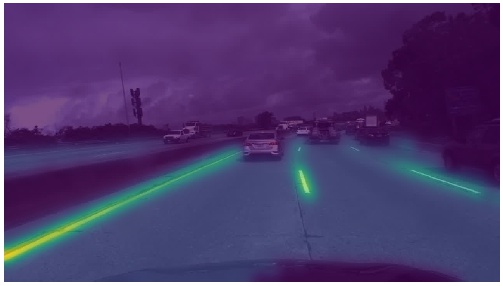}
  \caption*{(f) Left Lane Change}
\endminipage\hfill
\minipage{0.25\textwidth}
  \includegraphics[width=\linewidth,trim = {0 0.8cm 0 0.8cm},clip]{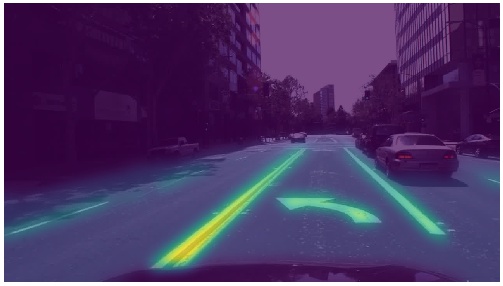}
  \caption*{(g) Left Lane Branch}
\endminipage\hfill
\minipage{0.25\textwidth}%
  \includegraphics[width=\linewidth,trim = {0 0.8cm 0 0.8cm},clip]{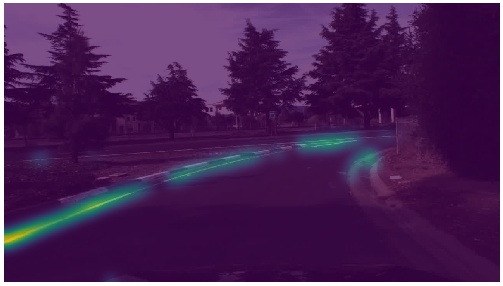}
  \caption*{(h) Merge}
\endminipage
\caption{Attention visualization from egocentric view. The first and second row show examples from Ego-Thing Graph and Ego-Stuff Graph, respectively. In (a)-(c), pedestrians intending to cross the street have significant influence on ego behavior when turning left, turning right and passing the crosswalk. The ego vehicle passes an intersection in (d) while paying attention to the moving car and bicycle in front of it. The figure (e) illustrates a left turn case when the heat map shows a high attention around the traffic light, which is green. In (f)-(h), lane markings show strong influences to ego's lane-related behaviors.}
\label{fig:visualization_egocentric_interaction}
\vspace{-8pt}
\end{figure*}

\begin{table}[t]
    \small
    \centering
        \begin{tabular}
            {@{}l l@{} c@{}}
            \toprule
              & Method &  Overall mAP \\
            \midrule
           \multirow{4}*{ \begin{tabular}{@{}c@{}}Different \\ Graphs\end{tabular}} &I3D~\cite{CarreiraCVPR2017} & 49.5 \\
            &Ego-Stuff Graph & 50.6 \\
            &Ego-Thing Graph & 50.8 \\
            &Ego-Thing Graph + Ego-Stuff Graph & \textbf{51.1} \\
            \midrule
             \multirow{2}*{\begin{tabular}{@{}c@{}}Spatial \\ Modeling \end{tabular}}&Appearance Relation  &  50.9 \\
            &Appearance + Spatial Relation & \textbf{51.1} \\
            \midrule
             \multirow{3}*{\begin{tabular}{@{}c@{}}Temporal \\ Modeling \end{tabular}}&Average & 50.0 \\
            &MLP & 50.9 \\
            &Max  &  \textbf{51.1} \\
            \bottomrule
        \end{tabular}
        \vspace{-3pt}
        \caption{Ablation Studies}
        \vspace{-15pt}
        \label{table:ablation_studies}
\end{table}

\subsection{Ablation Studies}\label{subsection:ablation_studies}

To provide a comprehensive understanding of the contributions from each module, we decompose our model into three components and conduct ablation studies using the \textbf{Goal-oriented} action recognition shown in Table \ref{table:ablation_studies}. 

\textbf{Comparison of Different Graphs.}
%
%
%
The first section of Table~\ref{table:ablation_studies} analyzes the influence of each graph to the tactical driver behavior recognition. 
The baseline is the I3D.
When ego-stuff graph or ego-thing graph is included, the results are boosted from 49.5\% to 50.6\% and 50.7\%, respectively.
If both graphs are trained jointly with the baseline model, we achieve the best performance 51.1\% on the \textbf{Goal-oriented} action recognition.
%
%

\textbf{Importance of Spatial Relation.} To investigate the effectiveness of spatial relation function in Eq. \ref{eq:4}, we conduct two experimental settings: using only the appearance relations, and embedding 3D spatial relation as an additional constraint. 
Without using the proposed 3D spatial relation, the performance decreases by 0.2\%, indicating the advantage of encoding spatial context.

\textbf{Variations of Temporal Modeling.} We analyze the impact of temporal modeling approaches. 
%
%
The best mAP -- 51.1\% is obtained by element-wise max pooling.
If we use element-wise averaging for the features from each time step, the model has a mAP of 50.0\%.
Our conjecture is that, for a 20-frame video clip, the key change takes place within a short duration. 
For example, in a \textit{Left Lane Change} behavior, the most noticeable moment is when the ego vehicle intersects the traffic lanes within a few frames. 
Temporal modeling using averaging features potentially degrades the distinguishable features, which will unavoidably result in information loss.
A multi-layer perceptron (MLP), which takes temporal ordering patterns into account, exceeds averaging pooling by 0.9\% but is 0.2\% lower than the best performance.
Our hypothesis is that significant change of interactive relations plays an more important role in recognizing tactical driver behavior than the ordering in time.

\subsection{Visualization}\label{subsection:visualization}
Apart from quantitative evaluation, we demonstrate interpretability of our method by the following two visualization strategies. 

\textbf{Attention Visualization from Egocentric View.} Given the learned affinity matrices in ego-thing graph and ego-stuff graph, we highlight those objects with strong connection to the ego node in Fig.~\ref{fig:visualization_egocentric_interaction}. 
%
%
The visualization results provide a strong proof that the proposed model captures the underlying interactions, which is essential for tactical driving behavior understanding.
Note that in the example shown in Fig~\ref{fig:visualization_egocentric_interaction} (e), the model captures the relation between the ego vehicle (turning left) and the traffic light (green light). 

\textbf{Attention Visualization from Top-view.} In addition to the interactions with ego, we can represent the complicated traffic scene in a graph structure as well. 
Fig.~\ref{fig:visualization_scene_representation2}(b) shows the visualized ego-thing graph  from the multi-head model for a scenario where the ego vehicle deviates for a parked truck. 
%
%
Each circle in the graph corresponds to a thing object in the frame and the ego vehicle is represented by a star.
The edge linking two nodes represents the interactive relation among them. 
%
%
We manually draw a top-view map Fig.~\ref{fig:visualization_scene_representation2}(c) to better represent the interactions based on spatial context.
%

\section{CONCLUSION}
In this paper, we propose a framework to model complicated interactions between driver and road users using graph convolution networks.
The proposed framework demonstrates favorable quantitative performance on the HDD dataset.
Qualitatively, we show the model can captures interactions between the ego vehicle and stuff objects, and the ego vehicle and thing objects. 
For future work, we plan to incorporate temporal modeling of thing objects into the proposed framework as it is an important cues in interaction modeling.
With that, the framework will enable tactical behavior anticipation and behavior modeling~\cite{Doshi_tacticalreview_itsc2011}, and potentially trajectory prediction of thing objects~\cite{Chandra_Traphic_CVPR2019}.  

\bibliographystyle{ieee}
\bibliography{reference}

\end{document}